\documentclass[letterpaper]{article} 
\usepackage{aaai2027}  
\usepackage[hyphens]{url}  
\usepackage{graphicx} 
\usepackage{natbib}  
\usepackage{caption} 
\usepackage{amsmath}  
\usepackage{amsfonts} 
\usepackage{algorithm}
\usepackage{algorithmic}
\usepackage{newfloat}
\usepackage{listings}
\DeclareCaptionStyle{ruled}{labelfont=normalfont,labelsep=colon,strut=off} 
\floatstyle{ruled}
\newfloat{listing}{tb}{lst}{}
\floatname{listing}{Listing}
\usepackage{booktabs}
\nocopyright
\title{Where Delay Compensation Belongs in a Decentralized Mean-Field-Game Swarm with a Control-Barrier Safety Filter}

\author{
    Jan Miko\l{}ajczyk\textsuperscript{\rm 1},
    Igor Rostkowski\textsuperscript{\rm 2},
    Bartosz Og\'orkiewicz\textsuperscript{\rm 1},
    Szymon Skierski\textsuperscript{\rm 1},
    Sylwia Opat\textsuperscript{\rm 1},
    Wojciech Rybak\textsuperscript{\rm 2},
    Miko\l{}aj Sczogiel\textsuperscript{\rm 2},
    Rados\l{}aw Pa\l{}ys\textsuperscript{\rm 3}, 
    Jan Gogulski\textsuperscript{\rm 4},
    Bartosz Ko\'scio\l{}ek\textsuperscript{\rm 1},
    Andrzej Pijanowski\textsuperscript{\rm 1},
    Szymon Kowalski\textsuperscript{\rm 1},
    Marcin Misiewicz\textsuperscript{\rm 4}
    }
\affiliations{
    \textsuperscript{\rm 1}University of Warsaw, Faculty of Physics and Faculty of Mathematics, Informatics and Mechanics\\
    \textsuperscript{\rm 2}University of Wroc\l{}aw, Faculty of Mathematics and Computer Science\\
    \textsuperscript{\rm 3}Wroc\l{}aw University of Science and Technology, Faculty of Information and Communication Technology\\
    \textsuperscript{\rm 4}Warsaw University of Technology, Faculty of Electronics and Information Technology
}
\begin{document}

\maketitle

\begin{abstract}
Large robot teams are often run as a plan-then-execute stack: a routing layer tells each agent where to go, and a safety layer keeps agents apart. We study a decentralized instance for drone swarms, a mean-field-game (MFG) routing field with a per-agent control-barrier-function (CBF) safety filter, when communication is limited by a shared TDMA channel. Agents know their neighbors only through noisy range and bearing measurements exchanged in round-robin slots, so the frame, and with it the age of every neighbor estimate, grows with the swarm size. We show in simulation that this delay reaches both layers but breaks only the safety layer. Without delay, swarms of up to 100 agents do not collide, even with the noise of each measurement held for a whole frame; with it, stale estimates at the CBF cause collisions beyond a frame period of 125 to 175\,ms, or 75 to 100\,ms at 100 agents with held noise, and the swarm size sets both where that limit lies and how steeply collisions rise beyond it. Stale estimates at the MFG distort the local density by which agents steer around congestion, but that density is an average over the crowd, which staleness makes noisier without biasing it, and throughput drops by at most a tenth. We then compensate the delay with lightweight on-board predictors, from constant-velocity fits to a small learned network, that estimate each neighbor's current position from its last broadcasts. A layer-split experiment shows where the prediction is needed: at the CBF it restores safety and extends the frame period the swarm tolerates, by a factor of 1.4 to 1.7 with an untrained constant-velocity fit and about 2.3 with the learned network when the measurement noise is redrawn at every step; with the noise held for a frame, the two extend it about equally, while at the MFG it only refines the congestion correction. A constant wider safety margin, the simplest alternative, makes collisions more frequent in a bottleneck. With the routing solved offline, the routing layer is robust to the delay because it sees the neighbors only through a crowd average, and compensation belongs in the safety filter, switched on only when the frame period exceeds the uncompensated limit, since below it the predictor's own error adds collisions.
\end{abstract}


\section{Introduction}
Multi-agent path finding (MAPF) plans collision-free paths for teams of agents, and a growing part of the field studies what happens when the plan meets execution: delays, limited communication, and agents that are not where the schedule expects them \citep{ma2017delay,honig2019adg,atzmon2020krobust,berndt2024receding}. For a large drone swarm heading to a shared goal, a decentralized alternative to a path per agent is a two-layer feedback controller, in which a routing layer gives each agent a velocity toward the goal and a reactive safety layer keeps agents apart \citep{wang2017safety,cui2023scalable,ballotta2025safe}. We study such a controller. Its routing layer is a mean-field game (MFG) \citep{lasry2007mfg,huang2006large}: an offline solve, whose cost does not depend on the number of agents, yields a congestion-aware velocity field defined at every position, so an agent that is delayed or displaced needs no replanning. Its safety layer is a control-barrier-function (CBF) filter \citep{ames2017cbfqp,wang2017safety}, which minimally modifies that velocity to keep each pair of agents apart. In MAPF terms, the routing field is the plan and the filter is its execution.

Both layers run on board, from each agent's estimate of where its neighbors are, and that estimate is late. Agents range each other over one shared channel. Under round-robin time-division multiple access (TDMA) every agent is heard once per frame and the frame has one slot per agent, so the age of a neighbor estimate grows in proportion to the swarm size: at 5\,ms per slot, 100 agents act on estimates up to half a second old. A swarm that grows thus loses the information its controller depends on, and simulated TDMA quadcopter swarms have been reported to collide for this reason beyond about twenty agents \citep{azem2022dynamic}.

We show how this delay acts on each layer. At the CBF, stale estimates cause collisions. A larger swarm is also a more crowded one, so we separate the two causes. Sweeping the slot duration at three fixed swarm sizes, so that each frame period is measured at three densities, and repeating the sweep with every swarm started at the same density, shows that no swarm collides up to a frame period of 125\,ms, that the last collision-free frame period falls from 175\,ms at 25 agents to 125\,ms at 100, and that collisions rise steeply beyond it, faster for larger swarms. Removing the delay while keeping the measurement noise leaves swarms of up to 100 agents free of collisions, even when the noise of each measurement is held for a whole frame, as on a real receiver; held noise does lower the limit, to 75 to 100\,ms at 100 agents. Staleness limits safety; the swarm size sets where that limit lies and how steeply collisions rise beyond it. At the MFG, stale estimates distort the local density with which agents steer around congestion. That density is an average over the crowd, which staleness makes noisier but barely biases, so the congestion correction, which is worth up to a third of the throughput, works about as well with stale estimates as with fresh ones; throughput drops by at most a tenth.

We then compensate the delay with on-board neighbor predictors. Each agent predicts a neighbor's current position from its last few broadcasts and their known ages; we compare constant-velocity fits, a rollout of the routing field, and a small learned network. None of these predictors is new, and prediction ahead of a barrier is established for CBFs \citep{molnar2023delay,ballotta2025safe}; they are our instruments. What is new here is the two-layer setting. Given an offline plan and a correction that sees the neighbors only through a density, one expects the routing layer to tolerate the delay better than the filter; our contribution is to measure how much better, and why. We switch the correction off to show what it is worth, and measure how staleness changes the density it reads. Giving the two layers different estimates then shows where the prediction is needed: at the CBF it restores safety, while at the MFG it only refines the congestion correction, making the swarm slightly safer and slightly slower. The controller should compensate the filter.

\paragraph{Contributions.}
\begin{itemize}
\item A staleness attribution. We show that TDMA staleness is what costs safety in an MFG+CBF swarm of 25 to 100 agents, using a slot-duration sweep at three swarm sizes, with fixed and with density-matched starts, and a delay-free control, which we repeat with the measurement noise held for a whole frame. Collisions start beyond a frame period of 125 to 175\,ms, a limit that falls with the swarm size, which also sets how steeply they rise. A delay budget, loosely analogous to $k$-robustness in MAPF, gives a scaling argument for that limit, consistent with its fall with the swarm size, but not a guarantee.
\item A layer split. The routing layer and the safety filter receive different neighbor estimates, with standard lightweight predictors as instruments. Prediction at the filter restores safety; prediction at the routing layer changes little. The routing layer's online congestion correction matters, worth up to a third of the throughput, but it uses the neighbors only through a local density, an average over the crowd that staleness barely biases. We found no prior work that varies which layer receives the compensated estimate.
\end{itemize}

This paper reports ongoing work, in simulation and with planar single-integrator agents. The next steps are to compute the routing online from local density estimates, in place of the offline solve, and to move to a hardware swarm; we welcome discussion of both.

\section{Related Work}

\paragraph{MAPF execution under delay.}
MAPF solvers compute conflict-free plans on a graph \citep{sharon2015cbs,stern2019mapf} and in continuous time \citep{andreychuk2022ccbs}, and executing those plans on real robots is a studied problem. MAPF-POST turns a plan into a schedule that respects kinematic constraints \citep{honig2016kinematic}, the Action Dependency Graph executes schedules robustly with reduced communication \citep{honig2019adg}, $k$-robust plans tolerate up to $k$ delays \citep{atzmon2020krobust}, receding-horizon re-ordering repairs the schedule when agents are delayed \citep{berndt2024receding}, and \citet{ma2017delay} plan with delay probabilities. Learned decentralized policies replace the plan with a reactive policy \citep{sartoretti2019primal} and learn with whom to communicate \citep{ma2021dcc}. For drone teams, decentralized trajectory planners have been made robust to communication delay below a known bound \citep{kondo2024rmader}. These approaches keep delay inside a discrete schedule or assume a bound on it. We study a continuous closed loop in which the delay grows with the team.

\paragraph{Mean-field routing.}
In a mean-field model an agent responds to the distribution of the others and not to each of them \citep{lasry2007mfg,huang2006large}, which makes the cost of planning independent of the team size \citep{elamvazhuthi2020meanfield} and reduces the communication a swarm needs \citep{shiri2020communication}. \citet{cui2023scalable} pair a learned mean-field control policy with artificial potential fields for collision avoidance, on real UAV swarms. Mean-field models do not constrain individual pairs, and density-level barriers \citep{wufung2025meanfield} constrain a continuum and not two physical vehicles, which is why we keep a per-agent filter on top of the routing field.

\paragraph{Barrier filters under delay.}
The CBF quadratic program is the standard minimally invasive safety filter \citep{ames2017cbfqp}, with decentralized pairwise certificates for multirobot teams \citep{wang2017safety} and learned variants that scale to large teams \citep{zhang2025gcbf}. Its guarantee assumes exact neighbor state. Robust variants tighten the constraint by a bound on the estimation error \citep{cosner2021measurement,dean2020guaranteeing,cheng2020safe}, and delay-aware variants handle sampled data and input delay, in particular by predicting the state across the delay \citep{singletary2020sampled,molnar2023delay}. Closest to us, \citet{ballotta2025safe} predict each neighbor's state from received data and its age before a learned distributed barrier, with a delay that grows with the local number of neighbors. Their nominal controller steers straight to the goal and uses no neighbor information, so only one layer can consume a prediction. In our controller both layers do, and the delay comes from a global schedule, so the slot duration changes staleness without changing the number or the placement of agents. \citet{azem2022dynamic} observe in simulation that a TDMA quadcopter swarm collides beyond 18 moving agents and respond with a new slot allocation; we keep the schedule and compensate in the controller.

\section{Controller}

We consider $N$ agents in a planar arena with walls. Agent $i$ has position $p_i\in\mathbb{R}^2$ and single-integrator dynamics $\dot p_i=u_i$ with speed limit $v_{\max}=8$\,m/s. It knows its own position and the static geometry exactly, and knows another agent $j$ only through an estimate $\hat p_{ij}$ of $p_j$. All agents head for one shared goal disc.

\paragraph{Offline mean-field game.}
The routing field solves an MFG with running cost $L=\tfrac12\lVert u\rVert^2+V_{\mathrm{goal}}(x)+V_{\mathrm{obs}}(x)+W(m)$, where $V_{\mathrm{goal}}$ pulls agents to the goal disc, $V_{\mathrm{obs}}$ is a smooth penalty on walls, and $W(m)=c_w m^2+\bar w\,(m-\bar m)_+^2$ penalizes density, more steeply above $\bar m=6.25$\,agents/m$^2$; $(\cdot)_+$ denotes $\max(\cdot,0)$. The goal pull is flat inside $0.75$ of the goal radius, so that arriving agents spread over the disc instead of packing at its center. The value function $\Phi$ and the density $m$ solve
\begin{align*}
-\partial_t\Phi-\nu\Delta\Phi+\tfrac12\lVert\nabla\Phi\rVert^2 &= V_{\mathrm{goal}}+V_{\mathrm{obs}}+W(m),\\
\partial_t m-\nu\Delta m-\nabla\!\cdot\!\left(m\,\nabla\Phi\right) &= 0,
\end{align*}
with $\Phi(T,\cdot)=0$ and $m(0,\cdot)=m_0$ of total mass $N$; the optimal velocity is $-\nabla\Phi$. We solve the system on a $0.2$\,m grid \citep{achdou2010meanfield}, with an explicit Lax-Friedrichs scheme for the value equation and an upwind finite-volume scheme for the density, by a damped fixed-point iteration related to fictitious play \citep{cardaliaguet2017learning,perrin2020fictitious}. The planned density $m^\star$ is the Ces\`aro average of the iterates after a burn-in, and $\Phi$ comes from a final value solve against $m^\star$. The viscosity $\nu=0.5$\,m$^2$/s regularizes the plan, while the agents themselves move without noise: without it the iteration cycles, and the density induced by the best response to $m^\star$ differs from $m^\star$ by a total variation of $0.95$, against $0.001$ with it. The smoothing is not small: after 10\,s the diffusion length $\sqrt{2\nu t}$ is about 3\,m, more than the gate width, so $m^\star$ is the equilibrium of a viscous relaxation of the dynamics the agents execute, not of those dynamics themselves. The viscosity is what makes the fixed point stable. The agents still benefit from the plan because they use it only through $-\nabla\Phi$, which carries the route and, with two gates, the split between them; the mismatch in density is left to the online correction below and to the filter. The solve is done once per layout and swarm size.

\paragraph{Routing layer.}
Online, agent $i$ follows the field and corrects for congestion that the plan did not anticipate:
\begin{align*}
u_{\mathrm{nom},i}={}&-\nabla\Phi(t,p_i)-c_{\mathrm{dens}}\nabla_{p_i}E_i,\\
E_i={}&c_w\big(\hat m_i-m^\star(t,p_i)\big)_+^2+\bar w\big(\hat m_i-\bar m\big)_+^2,
\end{align*}
where $\hat m_i$ is a Gaussian kernel density estimate, with bandwidth $0.6$\,m, over the estimated positions of the neighbors within $5$\,m. The correction pushes an agent out of regions where the local density exceeds the plan and vanishes otherwise. The neighbor estimates enter the routing layer through $\hat m_i$.

\paragraph{Safety filter.}
Separation is enforced by a per-agent CBF quadratic program in the decentralized pairwise form of \citet{wang2017safety}. For agent $i$ and neighbor $j$ the barrier is $h_{ij}=\lVert p_i-\hat p_{ij}\rVert^2-d_{\mathrm{safe}}^2$, with $d_{\mathrm{safe}}=0.70$\,m, a margin of $0.40$\,m above the collision distance $D_{\mathrm{coll}}=0.30$\,m of two airframes of radius $0.15$\,m. Each agent enforces its share $a_{ij}$ of the barrier condition, with $\hat n_{ij}$ the unit vector from $\hat p_{ij}$ to $p_i$ and $\gamma=20\,\mathrm{s}^{-1}$:
\[
\hat n_{ij}^{\top}u\;\ge\;b_{ij},\qquad b_{ij}=-\frac{a_{ij}\,\gamma\,h_{ij}}{2\lVert p_i-\hat p_{ij}\rVert}.
\]
With exact state the two shares of a pair sum to one.
They are biased toward the goal to keep traffic moving: the agent nearer the goal takes the larger
share, between $0.25$ and $0.75$, and the one behind yields.
Inside the safe set, where $h_{ij}<0$, the bias is dropped and both agents take one half, so that
recovery from a violation is symmetric.
The filter solves, over the eight nearest estimated neighbors $\mathcal{N}_i$ within $5$\,m,
\begin{align*}
u_i=\arg\min_{u}\;&\lVert u-u_{\mathrm{nom},i}\rVert^2\\
\text{s.t.}\;\;&\hat n_{ij}^{\top}u\ge b_{ij}-\delta_i,\quad j\in\mathcal{N}_i,\\
&\text{wall and boundary constraints},\\
&\lVert u\rVert_\infty\le v_{\max}/\sqrt{2}.
\end{align*}
The slack $\delta_i\ge0$ is the smallest uniform relaxation of the neighbor constraints that makes the program feasible, so it is zero whenever the hard problem is feasible; otherwise the filter returns a minimum-violation velocity and never an emergency stop. Wall constraints are never relaxed by a crowd. Because $u$ is two-dimensional, the program is solved exactly by vertex enumeration. With exact state and zero slack the filter keeps every constrained pair apart. The estimate $\hat p_{ij}$ enters the barrier value, the constraint direction, the share, and the choice of neighbors.

\section{TDMA Staleness and the Delay Budget}
Agents take turns on one channel. With one slot of duration $t_{\mathrm{slot}}$ per agent, the frame period is $T_{\mathrm{frame}}=N\,t_{\mathrm{slot}}$, and the age of the latest broadcast of agent $j$ is the sawtooth $\alpha_j(t)=(t-j\,t_{\mathrm{slot}})\bmod T_{\mathrm{frame}}$. At the slot of agent $j$, every other agent $i$ measures the relative vector $p_j-p_i$ in polar form, with Gaussian errors of standard deviation $\sigma_r=0.10$\,m in range and $\sigma_\theta=2^\circ$ in bearing, representing ultra-wideband ranging and a vision-based bearing. Without compensation, the agent adds this stale vector to its current position:
\[
\hat p_{ij}(t)=p_i(t)+\tilde r_{ij}\big(t-\alpha_j(t)\big),
\]
where $\tilde r_{ij}(s)$ is the noisy measurement of $p_j(s)-p_i(s)$. With $\Delta p_k$ the displacement of agent $k$ since the broadcast and $\varepsilon_{ij}$ the measurement error,
\[
\hat p_{ij}(t)-p_j(t)=\varepsilon_{ij}-\left(\Delta p_j-\Delta p_i\right).
\]
The staleness error is the relative displacement of the pair since the broadcast, of norm at most $v_{\mathrm{rel}}\,\alpha_j$ for a relative speed $v_{\mathrm{rel}}$. This models a single collision domain, as in a single-hop ultra-wideband network, and is a worst case relative to schedules with spatial reuse.

\paragraph{Delay budget.}
The margin gives a scaling argument for the delay, loosely analogous to $k$-robust MAPF \citep{atzmon2020krobust} but not a guarantee given in advance, since here the agents are on time and the information about them is late. The filter keeps the estimated separation above $d_{\mathrm{safe}}$, so a collision requires an estimate that is wrong by more than $d_{\mathrm{safe}}-D_{\mathrm{coll}}=0.40$\,m, and the controller tolerates estimates up to
\[
\alpha^\star=\frac{d_{\mathrm{safe}}-D_{\mathrm{coll}}}{v_{\mathrm{rel}}}
\]
old. At a relative speed of $0.8$\,m/s, for example, the budget is $0.5$\,s, the frame period of 100 agents at 5\,ms slots; pairs that close in on each other move several times faster, which is why the observed limit is far shorter. The relative speed of pairs that are already close is not known in advance, so the budget does not
say at which frame period collisions start.
It predicts what they depend on: chiefly the frame period $N\,t_{\mathrm{slot}}$, with the packing of
the agents entering through $v_{\mathrm{rel}}$, which grows as larger swarms close in denser groups.
Range noise is four standard deviations inside the margin, but that alone does not make it harmless. A single draw exceeds the margin with a probability of order $10^{-5}$, which is not rare at about $10^5$ neighbor measurements per second for $N=100$. It is harmless because each draw lasts one control step of 10\,ms, in which an agent moves at most 8\,cm, and that property comes from redrawing the noise at every step, a simplification we test in the Results by holding the noise for a whole frame.

\section{Delay-Compensating Predictors}
A predictor replaces the stale estimate by an estimate of the neighbor's \emph{current} position, computed on board from the neighbor's last broadcasts and their ages. No velocity is broadcast. We compare four:
\begin{itemize}
\item \textbf{cv}: constant-velocity extrapolation from the last two broadcasts. Differencing amplifies range noise into a velocity error of $\sqrt2\,\sigma_r/T_{\mathrm{frame}}$, which is $1.1$\,m/s at a 125\,ms frame.
\item \textbf{cv-ls}: a constant-velocity line fitted by least squares through the last four broadcasts, each corrected for the agent's own displacement since.
\item \textbf{field rollout}: the stale estimate integrated forward over its age along the routing
field $-\nabla\Phi$, on the assumption that the neighbor follows the shared routing policy.
It uses only the routing term: the congestion correction is left out, because the local density it
needs is defined at the observing agent's own position and not at a neighbor's unobserved one.
The field is read at the grid cell nearest the predicted position, one order coarser than the
interpolation the routing layer uses on its own position.
\item \textbf{learned}: a multilayer perceptron with two hidden layers of 128 units. Its input is the last four broadcasts, each as range, bearing, age, and the agent's own displacement since, and its output is the neighbor's current relative position. It is trained offline on windows pooled over swarm sizes from 25 to 200 and then frozen, so one network serves every pair and every $N$; the training range is wider than the 25 to 100 agents evaluated here, which the start chamber limits (Experimental Setup).
\end{itemize}
They are compared against four references: \textbf{real}, the uncompensated estimate; \textbf{noise-only}, the same measurements with the age set to zero, which is what perfect compensation would deliver; \textbf{exact positions}, the delay- and noise-free bound; and \textbf{margin}, the uncompensated estimate with a wider constant $d_{\mathrm{safe}}$, from 0.7 to 1.3\,m, the simplest alternative to a predictor; measurement-robust CBFs \citep{cosner2021measurement} instead scale the margin with the estimation error, which we leave to future work. In the \textbf{layer-split} configurations the routing layer and the safety filter receive different estimates, computed from the same measurements: prediction for the filter only, for the routing layer only, or for both.

\section{Experimental Setup}

\begin{figure*}[t]
\centering
\includegraphics[width=\textwidth]{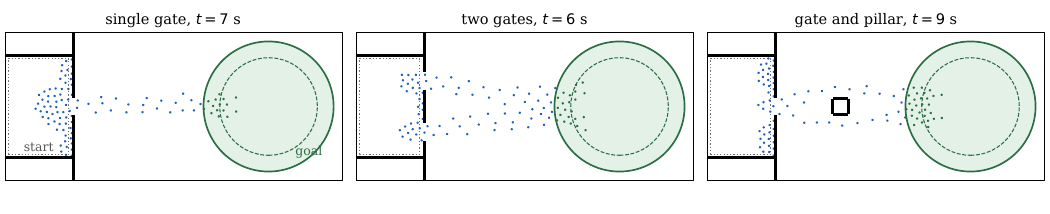}
\caption{The three layouts, $43.4\,\mathrm{m}\times19.1\,\mathrm{m}$, with a snapshot of a 100-agent swarm
(noise-only, one seed), agents drawn at their true size. Dotted: start chamber; the swarm crosses the
wall through the gates to the goal disc, whose goal pull is flat inside the dashed core. Green agents
have arrived.}
\label{fig:layouts}
\end{figure*}

All rollouts share one arena of $43.4\,\mathrm{m}\times19.1\,\mathrm{m}$, so that $N$ changes the
number of agents and the frame period but never the geometry.
A wall separates the source chamber, where agents start on a randomly jittered hexagonal lattice,
from the goal disc.
Three layouts set the routing problem (Figure~\ref{fig:layouts}): one centered gate of width $2.31$\,m, where routing is
trivial; two such gates, between which the swarm has to split; and the centered gate followed by a
square pillar of side $2$\,m in the flow.

The swarm size is $N\in\{25,50,100\}$, which at $t_{\mathrm{slot}}=5$\,ms gives frame
periods from $0.125$\,s to $0.5$\,s; beyond about 120 agents the start lattice is denser than
$d_{\mathrm{safe}}$, so collisions at the start would mask the effect of delay.
The staleness sweep varies $t_{\mathrm{slot}}$ at the three swarm sizes, on a grid chosen so that
they share the same eight frame periods from 75 to 300\,ms, which measures each frame period at
three different densities.
Because the start chamber is the same for every $N$, a second sweep starts the 25- and 50-agent
swarms at the 100-agent lattice spacing of 1.05\,m, which separates the start density from the
swarm size.
The staleness sweep uses the uncompensated estimate alone: it serves to attribute the collisions,
not to test compensation.

The closed loop runs at a control step of $10$\,ms over a horizon of three times the longest
start-to-goal path length at $1$\,m/s, with ten seeds per configuration; the seed sets the initial
positions and all noise.
Safety is measured by the fraction of agents that come within $D_{\mathrm{coll}}$ of another agent
at any time in the run; a run is collision-free when that fraction is zero.
A collision is recorded as a proximity event: it neither removes the agents involved nor changes
their dynamics, so agents that collide fly on and can still reach the goal, and safety and
throughput are measured independently.
Measuring it per agent, rather than counting pairs or violating time steps, keeps it comparable
across swarm sizes, which the staleness sweep requires.
Throughput is measured by the arrival rate at the goal in the bulk of the run,
$Q=0.8N/(t_{90}-t_{10})$ agents per second, where $t_x$ is the first time that $x\%$ of the agents
are inside the goal disc, so that neither the travel time of the first agents nor the last
stragglers enter it. $Q$ is reported only for configurations in which every run brings 90\% of the
agents to the goal; otherwise it is undefined and marked ---.
We report means over seeds; standard errors are given in the table captions.
Differences between arms at the same swarm size are paired on shared seeds, since a seed fixes the
same initial configuration; differences across swarm sizes are not, because a seed at one size has
no counterpart at another, and are reported as comparisons between independent seed sets.
The uncompensated cells at 125 to 175\,ms, where collisions set in, have thirty seeds. With ten seeds, a
configuration without any collision bounds the per-run collision probability at about $0.26$ with 95\%
confidence, and with thirty at $0.10$.

\section{Results}

\begin{figure}[t]
\centering
\includegraphics[width=\columnwidth]{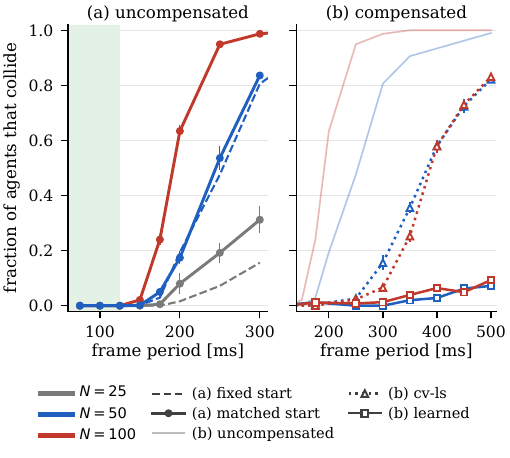}
\caption{Fraction of agents that collide against the TDMA frame period on the single gate, mean and
standard error over ten seeds (thirty at 125 to 175\,ms in panel a). (a) Uncompensated. Dashed: the
shared start chamber; solid with dots: every swarm starts at the 100-agent spacing, which for 100
agents is the shared chamber itself. Shaded: no collision in any run. (b) Compensated at 50 and 100
agents, learned (squares) and cv-ls (triangles, dotted), with the uncompensated curves faint for
reference.}
\label{fig:threshold}
\end{figure}

\paragraph{Staleness costs safety.}
Figure~\ref{fig:threshold} shows the fraction of agents that collide against the frame period, for
the uncompensated estimate. Up to 125\,ms no agent collides at any of the three swarm sizes, with
either start: 250 runs without a collision, pooled over the three swarm sizes, both starts and the
frame periods up to 125\,ms, which bounds the average per-run collision probability over these
configurations at 1.2\% with 95\% confidence (a single configuration at 125\,ms, with thirty runs, is
bounded at $0.10$). The last collision-free frame period is 175\,ms for 25 agents and 125\,ms for 50
and 100; at 150\,ms one run in thirty collides at 50 agents and half of the runs at 100, involving
0.1\% and 2.1\% of the agents, and at 250\,ms, 7\%, 48\% and 95\% of the agents collide.
Starting the smaller swarms at the 100-agent spacing raises their collisions, to 19\% and 54\% at
250\,ms, but does not close the gap to the 100-agent swarm, so the swarm size acts beyond the start
density. The first contacts happen in the first seconds, in the start chamber and at the gate, at
relative speeds that rise with the swarm size, from 4.0 to 5.1\,m/s at 200\,ms. This is the
dependence the delay budget expects: faster-closing swarms exhaust it sooner. The budget is
conservative, however. At 4.0 to 5.1\,m/s, a margin of 0.40\,m lasts 78 to 100\,ms, shorter than the
observed limit, because the budget takes the whole frame as the age of an estimate, while the age is a
sawtooth whose mean is half a frame; with per-step noise it is a lower bound on the collision-free
frame period, not an estimate of it. The budget leaves the noise out, and held noise shares the margin
and can bring the limit down to the budget (Held measurement noise below).
Without the delay no swarm collides, on any of the three layouts (noise-only in
Tables~\ref{tab:arms} and~\ref{tab:split}).

\paragraph{Compensation restores safety.}
Table~\ref{tab:arms} compares the predictors on the single gate at $t_{\mathrm{slot}}=5$\,ms, which
puts the 50- and 100-agent swarms well beyond the collision-free limit. The learned predictor keeps
the 50-agent swarm collision-free at a 250\,ms frame, where the uncompensated swarm loses half its
agents, and keeps 91\% of a 100-agent swarm collision-free at 500\,ms, where the uncompensated swarm
loses all of them. Its error on close neighbors is 0.020\,m$^2$ at 100 agents, against 0.24\,m$^2$
uncompensated and 0.011\,m$^2$ for the noise alone. cv-ls compensates at 50 agents but not at 100,
and cv, which differences two noisy broadcasts, does worse than no compensation. The field rollout
deadlocks: 59 to 93\% of the agents never reach the goal.
The learned predictor was trained on the single gate and transfers to the two-gate layout unchanged:
at 100 agents 8\% of the agents collide, against all of them uncompensated (Table~\ref{tab:split}).
Across frame periods (Figure~\ref{fig:threshold}), compensation extends the frame period that the
swarm tolerates, by a factor of 1.4 to 1.7 with cv-ls, which needs no training, and of 2.3 with
the learned predictor, with per-step noise. Uncompensated, at most 5\% of the agents collide up to 175\,ms at 50 agents and
150\,ms at 100; with cv-ls up to 250\,ms at both sizes, and with the learned predictor up to 400\,ms at
50 agents and 350\,ms at 100, with 7\% and 9\% at 500\,ms. Compensation does not reach zero, however: the learned predictor adds a
small error of its own, and leaves up to 1.2\% of the agents colliding at short frames, including
0.4\% at 125\,ms with 50 agents, where the uncompensated swarm has no collision at all. A controller should therefore compensate only above the
uncompensated limit.

\paragraph{Held measurement noise.}
Our runs redraw the measurement noise at every 10\,ms control step, whereas a receiver holds one sample
per broadcast. Because this could favor the delay-free control, we repeated the 100-agent runs on the
single gate with the noise of each broadcast drawn once and held until the sender's next broadcast.
The error itself is unchanged (the mean squared estimate error of noise-only is 0.011\,m$^2$ either
way); only its duration differs. Noise-only stays free of collisions at 125, 250 and 500\,ms (30 runs),
so held noise alone does not cause collisions, and the attribution to staleness stands. Held noise does
make stale estimates more dangerous: a held error is a bias that lasts a whole frame instead of a jitter
that the filter averages out, and it uses up part of the margin that staleness would otherwise have.
Uncompensated, 0.2\% of the agents collide at 75\,ms, 6.3\% at 100\,ms and 48\% at 125\,ms, so the limit
at 100 agents falls from 125\,ms to between 75 and 100\,ms. Compensation still extends it: with cv-ls
and with the learned predictor at most 5\% of the agents collide up to 175\,ms, and at 125\,ms cv-ls has
no collision where the uncompensated swarm has 48\%. The learned predictor, trained with per-step noise,
loses part of its advantage: it leaves 1.4\% of the agents colliding at 125\,ms and is ahead of cv-ls
only from 350\,ms (28\% against 48\%). Up to 250\,ms the untrained cv-ls is as safe as the network
(0\%, 1.7\% and 11\% against 1.4\%, 4.9\% and 9.2\% at 125, 175 and 250\,ms), so under held noise the
advantage of learning disappears, and a fit that needs no training suffices. The limits measured with per-step noise are therefore optimistic
for hardware, while the attribution and the gain from compensation carry over.

\begin{table}[t]
\centering
\small
\setlength{\tabcolsep}{3.5pt}
\begin{tabular}{lcccccc}
\toprule
 & \multicolumn{2}{c}{$N{=}25$, 125\,ms} & \multicolumn{2}{c}{$N{=}50$, 250\,ms} & \multicolumn{2}{c}{$N{=}100$, 500\,ms}\\
arm & coll. & $Q$ & coll. & $Q$ & coll. & $Q$\\
\midrule
noise-only  & 0    & 3.13 & 0    & 4.07 & 0    & 3.69\\
real        & 0    & 3.05 & 0.48 & 3.94 & 1.00 & 3.67\\
\midrule
cv          & 0    & 3.04 & 0.84 & 3.50 & 1.00 & 3.37\\
cv-ls       & 0    & 3.04 & 0.03 & 3.76 & 0.83 & 3.37\\
field rollout & 0.36 & --- & 0.92 & --- & 1.00 & ---\\
learned     & 0    & 3.06 & 0    & 3.66 & 0.09 & 2.73\\
\bottomrule
\end{tabular}
\caption{Single gate, $t_{\mathrm{slot}}=5$\,ms: fraction of agents that collide and throughput $Q$
in agents/s, mean of ten seeds (standard errors at most 0.06 and 0.08). The field rollout deadlocks,
so $Q$ is undefined. Collisions do not stop agents, so $Q$ of arms in which most agents collide is not
comparable with that of safe arms.}
\label{tab:arms}
\end{table}

\paragraph{The filter is where the prediction is needed.}
Table~\ref{tab:split} gives the layer split with two gates, where the swarm has to split between the
gates. Prediction at the filter alone gives nearly all the safety of prediction at both layers, with
either predictor and at both swarm sizes. Prediction at the routing layer alone leaves safety and
throughput where the uncompensated swarm has them. Added on top of a compensated filter, it makes
the 100-agent swarm slightly safer, from 11\% to 8\% of the agents colliding, and slightly slower,
from 4.27 to 4.00 agents/s.
This is not because the routing layer's use of the neighbors is minor. Switching its congestion
correction off ($c_{\mathrm{dens}}=0$), so that agents follow the offline plan and the filter alone,
costs a fifth of the noise-only throughput at 50 agents (5.23 against 6.53 agents/s) and more than a
third at 100 (3.16 against 4.97), and with the learned predictor it costs 17\% and 59\% (5.50 against
6.63, 1.63 against 4.00). Without it, the fraction of uncompensated agents that collide at 50 agents
rises from 35\% to 53\%. The correction spreads the crowd before the gates and so halves how often the
filter has to intervene (22\% of agent-steps against 40\% with noise-only at 50 agents).
The single gate shows the same split: prediction at the routing layer alone leaves 48\% of a 50-agent
swarm colliding, as uncompensated, while prediction at the filter alone leaves 1\%.

\begin{table}[t]
\centering
\small
\setlength{\tabcolsep}{4pt}
\begin{tabular}{llcccc}
\toprule
 & & \multicolumn{2}{c}{$N{=}50$, 250\,ms} & \multicolumn{2}{c}{$N{=}100$, 500\,ms}\\
routing & filter & coll. & $Q$ & coll. & $Q$\\
\midrule
real    & real    & 0.35  & 5.90 & 1.00 & 5.23\\
learned & real    & 0.35  & 5.97 & 1.00 & 5.19\\
real    & learned & 0.004 & 6.75 & 0.11 & 4.27\\
learned & learned & 0.004 & 6.63 & 0.08 & 4.00\\
\midrule
cv-ls   & real    & 0.35  & 5.99 & 1.00 & 5.29\\
real    & cv-ls   & 0.02  & 7.01 & 0.81 & 7.61\\
cv-ls   & cv-ls   & 0.03  & 7.20 & 0.83 & 7.95\\
\midrule
\multicolumn{2}{l}{noise-only} & 0 & 6.53 & 0 & 4.97\\
\multicolumn{2}{l}{exact positions} & 0 & 7.18 & 0 & 5.62\\
\bottomrule
\end{tabular}
\caption{Layer split with two gates, $t_{\mathrm{slot}}=5$\,ms: which estimate each layer receives.
Fraction of agents that collide and throughput $Q$ in agents/s, mean of ten paired seeds (standard
errors at most 0.07 and 0.15). Collisions do not stop agents, so $Q$ of arms in which most agents
collide, such as cv-ls in both layers, is not comparable with that of safe arms.}
\label{tab:split}
\end{table}

\paragraph{Throughput is not a casualty of staleness.}
The uncompensated swarm is about as fast as the noise-only one on the single gate (3.67 against 3.69
agents/s at 100 agents), and faster with two gates (5.23 against 4.97). Only with two gates at 50
agents does staleness cost throughput (5.90 against 6.53), and prediction at the filter recovers it.
Fast and unsafe go together: cv-ls at 100 agents with two gates is the fastest configuration in
Table~\ref{tab:split}, faster than exact positions, while 83\% of its agents collide. Agents that do
not see their neighbors in time do not yield to them, and since a collision in our simulation neither
removes an agent nor changes its dynamics, the agents that collide fly on and still count toward $Q$.

\section{Discussion}

\paragraph{Why the routing layer is robust to the delay.}
The routing layer consumes neighbor estimates in only one place, the online congestion correction.
That correction matters, as the $c_{\mathrm{dens}}=0$ runs show, but it reads the neighbors only
through a density averaged over the crowd within 5\,m, and a crowd that moves together keeps about
the same density in a stale snapshot. Staleness therefore makes the density noisier without biasing
it: at 100 agents with two gates its relative error grows from 8\% to 40\%, while inside crowds its
bias is $-3\%$, so the correction sees about the right excess over the plan and does its work as
well with stale estimates as with fresh ones. The offline plan, which carries the routing itself
and the split between the two gates, uses no neighbor estimates at all. This makes the routing layer
robust to communication delay, as long as the routing itself is solved offline and the neighbors
enter it only through a crowd average, in addition to having a planning cost independent of the swarm
size. Online routing would remove that protection (see Open questions). For this architecture, the
communication budget of a plan-then-execute stack is best spent on the execution layer. The safety filter, in contrast, acts on each pair, and
there an error of 44\,cm, the typical staleness error of a close neighbor at 100 agents and a 0.5\,s
frame, is the difference between a near miss and a collision.

\begin{figure}[t]
\centering
\includegraphics[width=\columnwidth]{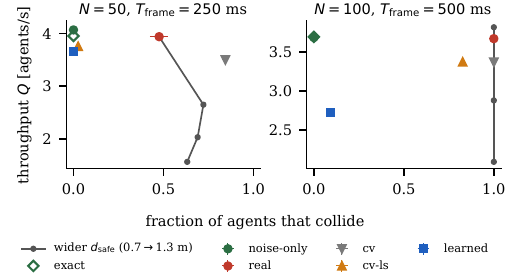}
\caption{Safety against throughput on the single gate at $t_{\mathrm{slot}}=5$\,ms, mean and standard
error over ten seeds. Line: the uncompensated estimate with a wider constant $d_{\mathrm{safe}}$,
0.7 to 1.3\,m (at 100 agents the widest margin leaves too few agents arriving to define $Q$). The
best configurations are at the top left.}
\label{fig:plane}
\end{figure}

\paragraph{Safety costs throughput.}
Compensation buys safety at a price. The learned predictor slows the 100-agent swarm by about a
quarter, on the single gate (2.73 against 3.70 agents/s with exact positions) and with two gates
(4.00 against 5.62), and its filter intervenes more often (59\% of agent-steps against 51\% with exact
positions). Throughput must therefore be read together with safety: an arm that is fast and unsafe is
not better than one that is slow and safe, and each configuration is a point in that plane rather than
a ranking on one axis (Figure~\ref{fig:plane}).

A constant wider safety margin, the simplest alternative to a predictor, does not buy safety in this setting.
Widening $d_{\mathrm{safe}}$ from 0.7 to 0.9, 1.1 and 1.3\,m raises the fraction of a 50-agent swarm
that collides from 48\% to 72\%, 69\% and 63\%, and cuts throughput by a third to three fifths; at 100
agents every margin leaves all agents colliding. A crowd in front of a 2.31\,m gate cannot keep the
wider spacing, so the filter has to relax its constraints, on 7\% and 12\% of agent-steps at 0.9 and
1.3\,m against 1\% at 0.7\,m, and the relaxation falls where the crowd is densest. A predictor removes
the error where it arises; a margin only asks for room that a bottleneck does not have.

\paragraph{What the threshold means for a TDMA swarm.}
The collision-free limit, 125\,ms at 100 agents and up to 175\,ms for smaller swarms, is a design rule for the
schedule: at 100 agents it requires slots of at most 1.25\,ms (0.75 to 1\,ms with held noise), and at 5\,ms per slot it allows 25
agents. Compensation at the filter extends it to 250\,ms with cv-ls and to 350 to 400\,ms with the learned
predictor, at a few percent of agents colliding, which at the same slot length allows a swarm 1.4 to
2.3 times as large on the same channel, at a throughput cost that grows with the frame period. With
held noise both predictors reach about 175\,ms at 100 agents, so the factor of 2.3 is specific to
the learned predictor under per-step noise.

\paragraph{Limitations.}
The field rollout deadlocks because it moves each neighbor along the routing field at up to the speed
limit, while congestion and the filter hold real agents to about half of that, so every agent yields
to neighbors that are not there; a model-based predictor without interaction terms is worse than none.
The planner's walls are a soft penalty that costs less to cross than a detour to a gate, so in front
of the wall between the two gates and in front of the pillar the field points into the obstacle, and
three to four agents per run, the same in every arm, stop there. On the pillar layout, exact
positions themselves deadlock more often than noisy ones: with exact state, 36\% and 27\% of 50- and
100-agent swarms never reach the goal, against 8\% and 4\% with noise, which suggests that the
measurement noise breaks symmetric standoffs at the filter. The learned predictor keeps the pillar
layout nearly collision-free (none at 50 agents, 8\% at 100) but is slowest there: at 50 agents it reaches half of the noise-only throughput (0.67 against 1.39
agents/s), and at 100 agents only 3 of 10 runs bring 90\% of the agents to the goal, so $Q$ is
undefined. The uncompensated swarm is again the fastest.
The learned predictor was trained on an earlier configuration of the controller and is used here
unchanged. Because a collision does not end a trajectory, the fraction of agents that collide
beyond the limit counts proximity events in a swarm that keeps flying, where a real swarm would
lose agents after the first few: where collisions start is a physical result, but how steeply they
rise beyond that point is partly a property of the simulator, and we read it as an indicator rather
than a prediction of losses. Runs on different GPUs are statistically equivalent but not bit-identical, because the
closed loop amplifies floating-point differences, so all paired comparisons are within one machine.

\paragraph{Scope.}
The results are for one MFG+CBF controller, planar single-integrator agents sized as five-inch
quadrotors, and one simulator.
The main runs redraw the measurement noise at every control step, where a receiver holds one sample
per frame; held noise, which we tested at 100 agents on the single gate only, leaves the attribution
intact but lowers the limits.
The routing field is solved offline for a known layout.
The learned predictor, behind the largest gain from compensation, was trained on an earlier
configuration of the controller, so it is an instrument rather than a validated component; cv-ls,
which needs no training, already reaches 250\,ms.

\paragraph{Open questions.}
\begin{itemize}
\item \emph{Online routing.} Routing computed online from local density estimates would consume
neighbor estimates far more heavily than the congestion correction does. Does the delay then reach
the routing layer, and does compensation belong there as well?
\item \emph{Impermeable walls.} Can the planner keep hard walls and still reach a stable fixed point,
and does that free the agents trapped in front of obstacles?
\item \emph{A predictor for this controller.} How much of the learned predictor's residual
collisions and throughput cost is due to its training on another configuration?
\item \emph{Hardware.} Do the limit and its extension survive real ultra-wideband ranging, with
errors correlated across frames, outliers and packet loss?
\end{itemize}

\section{Conclusion}
Under a shared TDMA channel, the age of every neighbor estimate in an MFG+CBF swarm grows with the
swarm size. In simulation, this staleness breaks the safety filter beyond a frame period of 125 to
175\,ms, depending on the swarm size and earlier when the measurement noise is held for a frame, and leaves the routing layer almost untouched, because the
offline plan uses no neighbor estimates and the online correction uses them only through a crowd
average. Prediction at the filter extends the frame period the swarm tolerates by a factor of 1.4 to 2.3 with per-step noise, and with held noise an untrained constant-velocity fit does as well as the learned predictor, while a constant wider
safety margin makes a bottleneck less safe. Whether this division of labor survives online routing
and real radios is open, and it is what we would most like to discuss at the workshop.

\bibliography{aaai2027}

@article{lasry2007mfg,
  author  = "Lasry, Jean-Michel and Lions, Pierre-Louis",
  title   = "Mean Field Games",
  journal = "Japanese Journal of Mathematics",
  volume  = 2, number = 1, pages = "229-260", year = 2007,
}

@article{huang2006large,
  author  = "Huang, Minyi and Malham{\'e}, Roland P. and Caines, Peter E.",
  title   = "Large Population Stochastic Dynamic Games: Closed-Loop {McKean-Vlasov} Systems and the {Nash} Certainty Equivalence Principle",
  journal = "Communications in Information and Systems",
  volume  = 6, number = 3, pages = "221-252", year = 2006,
}

@article{elamvazhuthi2020meanfield,
  author  = "Elamvazhuthi, Karthik and Berman, Spring",
  title   = "Mean-Field Models in Swarm Robotics: A Survey",
  journal = "Bioinspiration \& Biomimetics",
  volume  = 15, number = 1, pages = "015001", year = 2020,
}

@article{cardaliaguet2017learning,
  author  = "Cardaliaguet, Pierre and Hadikhanloo, Saeed",
  title   = "Learning in Mean Field Games: The Fictitious Play",
  journal = "ESAIM: Control, Optimisation and Calculus of Variations",
  volume  = 23, number = 2, pages = "569-591", year = 2017,
}

@inproceedings{perrin2020fictitious,
  author    = "Perrin, Sarah and P{\'e}rolat, Julien and Lauri{\`e}re, Mathieu and Geist, Matthieu and Elie, Romuald and Pietquin, Olivier",
  title     = "Fictitious Play for Mean Field Games: Continuous Time Analysis and Applications",
  booktitle = "Advances in Neural Information Processing Systems (NeurIPS)",
  volume    = 33, year = 2020,
}

@article{achdou2010meanfield,
  author  = "Achdou, Yves and Capuzzo-Dolcetta, Italo",
  title   = "Mean Field Games: Numerical Methods",
  journal = "SIAM Journal on Numerical Analysis",
  volume  = 48, number = 3, pages = "1136-1162", year = 2010,
}

@inproceedings{cui2023scalable,
  author    = "Cui, Kai and Li, Mengguang and Fabian, Christian and Koeppl, Heinz",
  title     = "Scalable Task-Driven Robotic Swarm Control via Collision Avoidance and Learning Mean-Field Control",
  booktitle = "Proceedings of the IEEE International Conference on Robotics and Automation (ICRA)",
  year      = 2023,
  doi       = "10.1109/ICRA48891.2023.10161498",
}

@article{shiri2020communication,
  author  = "Shiri, Hamid and Park, Jihong and Bennis, Mehdi",
  title   = "Communication-Efficient Massive {UAV} Online Path Control: Federated Learning Meets Mean-Field Game Theory",
  journal = "IEEE Transactions on Communications",
  volume  = 68, year = 2020,
}

@inproceedings{wufung2025meanfield,
  author    = "Wu Fung, Samy and Nurbekyan, Levon",
  title     = "Mean-Field Control Barrier Functions: A Framework for Real-Time Swarm Control",
  booktitle = "Proceedings of the American Control Conference (ACC)",
  year      = 2025,
}

@article{ames2017cbfqp,
  author  = "Ames, Aaron D. and Xu, Xiangru and Grizzle, Jessy W. and Tabuada, Paulo",
  title   = "Control Barrier Function Based Quadratic Programs for Safety Critical Systems",
  journal = "IEEE Transactions on Automatic Control",
  volume  = 62, number = 8, pages = "3861-3876", year = 2017,
}

@article{wang2017safety,
  author  = "Wang, Li and Ames, Aaron D. and Egerstedt, Magnus",
  title   = "Safety Barrier Certificates for Collisions-Free Multirobot Systems",
  journal = "IEEE Transactions on Robotics",
  volume  = 33, number = 3, pages = "661-674", year = 2017,
}

@article{zhang2025gcbf,
  author  = "Zhang, Songyuan and So, Oswin and Garg, Kunal and Fan, Chuchu",
  title   = "{GCBF+}: A Neural Graph Control Barrier Function Framework for Distributed Safe Multi-Agent Control",
  journal = "IEEE Transactions on Robotics",
  volume  = 41, year = 2025,
  doi     = "10.1109/TRO.2025.3530348",
}

@inproceedings{cosner2021measurement,
  author    = "Cosner, Ryan K. and Singletary, Andrew W. and Taylor, Andrew J. and Molnar, Tamas G. and Bouman, Katherine L. and Ames, Aaron D.",
  title     = "Measurement-Robust Control Barrier Functions: Certainty in Safety with Uncertainty in State",
  booktitle = "Proceedings of the IEEE/RSJ International Conference on Intelligent Robots and Systems (IROS)",
  pages     = "6286-6291", year = 2021,
}

@inproceedings{dean2020guaranteeing,
  author    = "Dean, Sarah and Taylor, Andrew J. and Cosner, Ryan K. and Recht, Benjamin and Ames, Aaron D.",
  title     = "Guaranteeing Safety of Learned Perception Modules via Measurement-Robust Control Barrier Functions",
  booktitle = "Proceedings of the Conference on Robot Learning (CoRL)",
  series    = "Proceedings of Machine Learning Research", volume = 155,
  pages     = "654-670", year = 2020,
}

@inproceedings{cheng2020safe,
  author    = "Cheng, Richard and Khojasteh, Mohammad Javad and Ames, Aaron D. and Burdick, Joel W.",
  title     = "Safe Multi-Agent Interaction through Robust Control Barrier Functions with Learned Uncertainties",
  booktitle = "Proceedings of the 59th IEEE Conference on Decision and Control (CDC)",
  pages     = "777-783", year = 2020,
}

@inproceedings{singletary2020sampled,
  author    = "Singletary, Andrew and Chen, Yuxiao and Ames, Aaron D.",
  title     = "Control Barrier Functions for Sampled-Data Systems with Input Delays",
  booktitle = "Proceedings of the 59th IEEE Conference on Decision and Control (CDC)",
  pages     = "804-809", year = 2020,
}

@article{molnar2023delay,
  author  = "Molnar, Tamas G. and Kiss, Adam K. and Ames, Aaron D. and Orosz, Gabor",
  title   = "Safety-Critical Control with Input Delay in Dynamic Environment",
  journal = "IEEE Transactions on Control Systems Technology",
  volume  = 31, number = 4, pages = "1507-1520", year = 2023,
}

@article{ballotta2025safe,
  author  = "Ballotta, Luca and Talak, Rajat",
  title   = "Safe Distributed Control of Multi-Robot Systems with Communication Delays",
  journal = "IEEE Transactions on Vehicular Technology",
  volume  = 74, number = 7, pages = "10137-10150", year = 2025,
}

@inproceedings{azem2022dynamic,
  author    = "Azem, Sharif and Tahir, Anam and Koeppl, Heinz",
  title     = "Dynamic Time Slot Allocation Algorithm for Quadcopter Swarms",
  booktitle = "Proceedings of the 19th IEEE Annual Consumer Communications \& Networking Conference (CCNC)",
  pages     = "1-6", year = 2022,
}

@article{kondo2024rmader,
  author  = "Kondo, Kota and Figueroa, Reinaldo and Rached, Juan and Tordesillas, Jesus and Lusk, Parker C. and How, Jonathan P.",
  title   = "Robust {MADER}: Decentralized Multiagent Trajectory Planner Robust to Communication Delay in Dynamic Environments",
  journal = "IEEE Robotics and Automation Letters",
  volume  = 9, number = 2, pages = "1476-1483", year = 2024,
}

@inproceedings{stern2019mapf,
  author    = "Stern, Roni and Sturtevant, Nathan R. and Felner, Ariel and Koenig, Sven and Ma, Hang and Walker, Thayne T. and Li, Jiaoyang and Atzmon, Dor and Cohen, Liron and Kumar, T. K. Satish and Boyarski, Eli and Bart{\'a}k, Roman",
  title     = "Multi-Agent Pathfinding: Definitions, Variants, and Benchmarks",
  booktitle = "Proceedings of the 12th International Symposium on Combinatorial Search (SoCS)",
  pages     = "151-158", year = 2019,
}

@article{sharon2015cbs,
  author  = "Sharon, Guni and Stern, Roni and Felner, Ariel and Sturtevant, Nathan R.",
  title   = "Conflict-Based Search for Optimal Multi-Agent Pathfinding",
  journal = "Artificial Intelligence", volume = 219, pages = "40-66", year = 2015,
}

@article{andreychuk2022ccbs,
  author  = "Andreychuk, Anton and Yakovlev, Konstantin and Surynek, Pavel and Atzmon, Dor and Stern, Roni",
  title   = "Multi-Agent Pathfinding with Continuous Time",
  journal = "Artificial Intelligence",
  volume  = 305, pages = "103662", year = 2022,
}

@article{sartoretti2019primal,
  author  = "Sartoretti, Guillaume and Kerr, Justin and Shi, Yunfei and Wagner, Glenn and Kumar, T. K. Satish and Koenig, Sven and Choset, Howie",
  title   = "{PRIMAL}: Pathfinding via Reinforcement and Imitation Multi-Agent Learning",
  journal = "IEEE Robotics and Automation Letters",
  volume  = 4, year = 2019,
}

@misc{ma2021dcc,
  author        = "Ma, Ziyuan and Luo, Yudong and Pan, Jia",
  title         = "Learning Selective Communication for Multi-Agent Path Finding",
  year          = 2021,
  eprint        = "2109.05413", archivePrefix = "arXiv",
}

@inproceedings{ma2017delay,
  author    = "Ma, Hang and Kumar, T. K. Satish and Koenig, Sven",
  title     = "Multi-Agent Path Finding with Delay Probabilities",
  booktitle = "Proceedings of the AAAI Conference on Artificial Intelligence",
  volume    = 31, pages = "3605-3612", year = 2017,
}

@inproceedings{honig2016kinematic,
  author    = "H{\"o}nig, Wolfgang and Kumar, T. K. Satish and Cohen, Liron and Ma, Hang and Xu, Hong and Ayanian, Nora and Koenig, Sven",
  title     = "Multi-Agent Path Finding with Kinematic Constraints",
  booktitle = "Proceedings of the International Conference on Automated Planning and Scheduling (ICAPS)",
  volume    = 26, pages = "477-485", year = 2016,
}

@article{honig2019adg,
  author  = "H{\"o}nig, Wolfgang and Kiesel, Scott and Tinka, Andrew and Durham, Joseph W. and Ayanian, Nora",
  title   = "Persistent and Robust Execution of {MAPF} Schedules in Warehouses",
  journal = "IEEE Robotics and Automation Letters",
  volume  = 4, number = 2, pages = "1125-1131", year = 2019,
}

@article{atzmon2020krobust,
  author  = "Atzmon, Dor and Stern, Roni and Felner, Ariel and Wagner, Glenn and Bart{\'a}k, Roman and Zhou, Neng-Fa",
  title   = "Robust Multi-Agent Path Finding and Executing",
  journal = "Journal of Artificial Intelligence Research",
  volume  = 67, pages = "549-579", year = 2020,
}

@article{berndt2024receding,
  author  = "Berndt, Alex and van Duijkeren, Niels and Palmieri, Luigi and Kleiner, Alexander and Keviczky, Tam{\'a}s",
  title   = "Receding Horizon Re-Ordering of Multi-Agent Execution Schedules",
  journal = "IEEE Transactions on Robotics",
  volume  = 40, pages = "1356-1372", year = 2024,
}

\end{document}